\pdfoutput=1

\documentclass[11pt]{article}

\usepackage{ACL2023}

\usepackage{times}
\usepackage{latexsym}

\usepackage[T1]{fontenc}

\usepackage[utf8]{inputenc}

\usepackage{microtype}
\usepackage{graphicx}
\usepackage{inconsolata}

\usepackage{CJKutf8}
\usepackage{algorithm}
\usepackage{algorithmic}
\usepackage{booktabs}
\usepackage{hyperref}
\usepackage{amsmath}
\usepackage{bm}
\usepackage{amssymb}
\usepackage{pifont}
\newcommand{\cmark}{\ding{51}}%
\newcommand{\xmark}{\ding{55}}%
\newcommand{\dataset}{MidMed}

\title{MidMed: Towards Mixed-Type Dialogues for Medical Consultation}

\author{
 Xiaoming Shi\textsuperscript{\rm 1}\thanks{\quad Equal contribution.}\textit{ }\textit{ }, Zeming Liu\textsuperscript{\rm 2}\footnotemark[1]\textit{ }\textit{ }, Chuan Wang\textsuperscript{3}, Haitao Leng\textsuperscript{4}, Kui Xue\textsuperscript{1}, \\
 \textbf{Xiaofan Zhang\textsuperscript{1}, Shaoting Zhang\textsuperscript{1}\thanks{\quad Corresponding author: Shaoting Zhang.}}\\
 \textsuperscript{1} Shanghai Artificical Intelligence Laboratory, Shanghai, China \\
 \textsuperscript{2} Research Center for Social Computing and Information Retrieval, HIT, Harbin, China \\
 \textsuperscript{3} State Key Laboratory of Information Security, IIE, CAS, Beijing, China \\
 \textsuperscript{4} MMU KuaiShou Inc., Hangzhou, China \\
 {\tt \{shixiaoming, xuekui, zhangxiaofan, shaotingzhang\}@pjlab.org.cn} \\ 
 {\tt zmliu@ir.hit.edu.cn; wangchuan@iie.ac.cn; lenghaitao@kuaishou.com}
}

\newcommand{\chuhao}{\fontsize{9pt}{\baselineskip}\selectfont}

\begin{document}
\maketitle
\begin{abstract}
Most medical dialogue systems assume that patients have clear goals (medicine querying, surgical operation querying, etc.) before medical consultation. 
However, in many real scenarios, due to the lack of medical knowledge, it is usually difficult for patients to determine clear goals with all necessary slots.
In this paper, we identify this challenge as how to construct medical consultation dialogue systems to help patients clarify their goals.
To mitigate this challenge, we propose a novel task and create a human-to-human mixed-type medical consultation dialogue corpus, termed MidMed~\footnote{MidMed is publicly available at https://github.com/xmshi-trio/MidMed}, covering five dialogue types: task-oriented dialogue for diagnosis, recommendation, knowledge-grounded dialogue, QA, and chitchat. 
MidMed covers four departments (otorhinolaryngology, ophthalmology, skin, and digestive system), with 8,175 dialogues.
Furthermore, we build baselines on MidMed and propose an instruction-guiding medical dialogue generation framework, termed InsMed, to address this task. 
Experimental results show the effectiveness of InsMed.
\end{abstract}

\section{Introduction}
Current medical dialogue systems~\citep{xu2019end,liao2020task,zeng-etal-2020-meddialog,liu2022meddg} mainly focus on diagnosis by obtaining symptoms and then making diagnosis automatically.
These dialogue systems have shown significant potential and alluring technological value to simplify diagnostic procedures~\citep{semigran2015evaluation}.
Previous works assume that patients have explicit goals (medicine querying, surgical operation querying, etc.), and perform in the way of task-oriented dialogue to accomplish patients' goals.

\begin{figure*}[t]
	\small
	\centering
	\includegraphics[width=\linewidth]{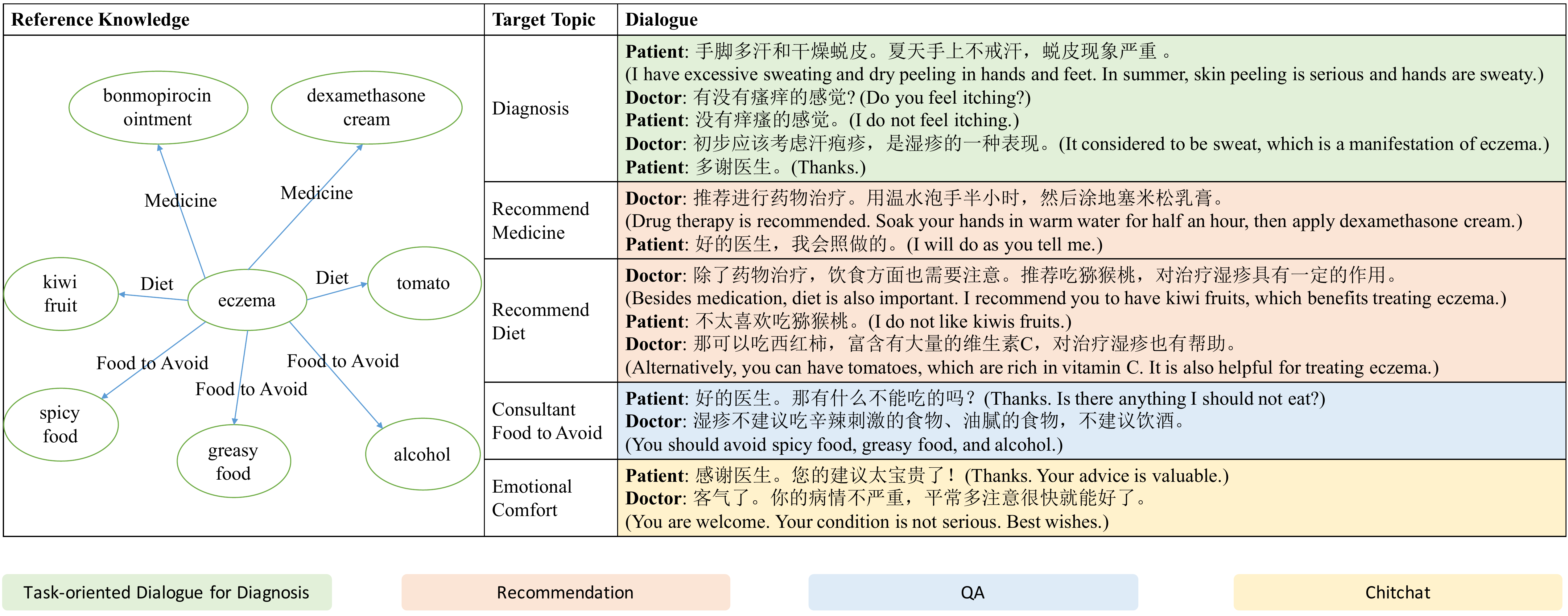}\\
	\caption{An example of MidMed.}
	\label{figure:sample}
\end{figure*}

However, explicit patient goals are usually unavailable in real-world scenarios.
For example, a patient wants to consult about his itchy skin but lacks medical knowledge.
Thus, it is difficult for the patient to decide which slots (e.g. medicine or a surgical operation) are needed.
To figure out explicit patient goals, medical consultation services are needed, which provide advice of treatment, medicine, food, etc., as shown in Figure~\ref{figure:sample}.
However, those medical consultation services are under explored in previous works.

To facilitate the study of medical consultation, we construct a new human-to-human \textbf{mi}xed-type \textbf{d}ialogue dataset for \textbf{med}ical consultation (MidMed), covering five dialogue types: task-oriented dialogue for diagnosis, knowledge-grounded dialogue, QA, recommendation, and chitchat.
MidMed is constructed by revising dialogues of MedDialog (a human-to-human medical diagnosis dialogue dataset)~\citep{zeng-etal-2020-meddialog}.
As shown in Figure~\ref{figure:sample}, a patient queries about “sweaty hands”, and has no explicit goal for medicine or a surgical operation. 
In the scenario, the doctor first collects the symptoms and makes a diagnosis.
To help clarify the patient's goal, the doctor further recommends medicine and food, replies for foods to avoid, and gives emotional comfort.
Through the consultation, the patient determines to apply ``dexamethasone cream'' and have more ``tomatoes''.
Finally, MidMed is obtained, containing 8,175 dialogues and 98,000 utterances, with at least three dialogue types in each dialogue.

To promote research on medical consultation dialogue systems, we conduct benchmarking experiments on MidMed for end-to-end dialogue generation.
Furthermore, to generate informative and relevant responses with dialogue topic sequences, inspired by~\citet{schick2020exploiting,wei2021finetuned}, we present an \textbf{ins}truction-guiding \textbf{me}dical \textbf{d}ialogue generation framework (InsMed) to handle mixed-type dialogues.
InsMed is composed of a dialogue topic selection, a reference knowledge selection, and an instruction-based response generation module.
Specifically, the topic selection module and the reference knowledge selection module are designed to pick suitable dialogue topics and reference knowledge for generating responses, respectively.
Then, dialogue topics and reference knowledge are converted to instructions in natural language with well-designed templates.
For example, an instruction is ``\textit{In the next utterance, the doctor will recommend a diet. The recommended diet is fruits and vegetables}''.
These instructions are concatenated with dialogue context as the input to generation models.

This work makes the following contributions:
\begin{itemize}
    \item We identify a new challenge, that is, in many real-world scenarios, it is usually difficult for patients to have clear goals before medical consultations.
    \item To mitigate this challenge, we propose a novel task, medical consultation over mixed-type dialogue, and collect a new Chinese human-to-human mixed-type dialogue dataset, in which each session has rich variability of dialogue types with natural topic transitions.
    \item We build baselines on MidMed and propose an instruction-guiding response generation framework InsMed to address this task. Experimental results show the effectiveness of InsMed.
\end{itemize}

\begin{table*}[t]
		\centering
		\small
		\begin{tabular}{ p{4.65cm} c  c c} 
			\toprule
			Datasets  & Mixed-type & Medical & Dialogue Types\\ 
			\midrule
                MZ~\citep{wei2018task} & \xmark & \cmark & Task-oriented dialogue for diagnosis \\
                DX~\citep{xu2019end} & \xmark & \cmark & Task-oriented dialogue for diagnosis \\
                CMDD~\citep{lin2019enhancing} & \xmark & \cmark & Task-oriented dialogue for diagnosis \\
                MedDG~\citep{liu2022meddg} & \xmark & \cmark & Task-oriented dialogue for diagnosis \\
                MedDialog~\citep{zeng-etal-2020-meddialog} & \xmark & \cmark & Task-oriented dialogue for diagnosis \\
                DialoAMC~\citep{chen2022benchmark} & \xmark & \cmark & Task-oriented dialogue for diagnosis \\\midrule
                DuRecDial \citep{liu-etal-2020} & \cmark & \xmark & Rec., chitchat, QA, task-oriented dialogue\\
			DodecaDialogue\citep{Kurt2020} & \cmark &\xmark & Know., chitchat, QA, empathetic dialogue, image chat \\
			BlendedSkillTalk\citep{smith-etal-2020} & \cmark &\xmark & Know., empathetic dialogue,chitchat\\
			ACCENTOR\citep{Sun2021AddingCT} & \cmark &\xmark & Chitchat,task-oriented dialogue\\
                DuRecDial 2.0 \citep{liu-etal-2021-du} & \cmark & \xmark & Rec., chitchat, QA, task-oriented dialogue\\
			SalesBot\citep{chiu-etal-2022} & \cmark &\xmark & Chitchat,task-oriented dialogue\\
			DuClarifyDial\cite{liu-etal-2022} & \cmark &\xmark & Rec., know. chitchat, QA, task-oriented dialogue\\\midrule
                MidMed (Ours) & \cmark &\cmark & Rec., chitchat, know., QA, diagnosis-oriented dialogue\\
        \bottomrule
		\end{tabular}
            \caption{Comparison of MidMed with other datasets. ``know.'',  and ``rec.'' stand for knowledge-grounded dialogue, and conversational recommendation, respectively.}
		\label{tab:mixed_type_datasets}
\end{table*}

\section{Related Work}
\subsection{Dialogue Systems for Diagnosis}
There has been growing research interest in developing dialogue systems for automatic diagnosis.
These dialogue systems aim to assist doctors in pre-collecting symptoms and patient information and then give patients diagnoses in time.
These works are divided into two categories, the pipeline manner, and the end-to-end manner.
\citet{wei2018task,xu2019end,lin2019enhancing,wang2021online,liu2022meddg} break the systems into natural language understanding, dialogue management, and natural language generation, in a pipeline manner.
Then, these three modules are trained with respective annotated data and feed their output to the next module.
Meanwhile, \citet{zeng-etal-2020-meddialog} tries to build an end-to-end model on large-scale unannotated medical dialogue data.
Compared with the pipeline manner, the end-to-end manner has no requirement for the annotated dataset but has no supervision for the intermediate state.

In addition to methods, many datasets are also publicly available.
The medical dialogue datasets are listed in Table~\ref{tab:mixed_type_datasets}.
Among them, MZ~\citep{wei2018task}, DX~\citep{xu2019end}, CMDD~\citep{lin2019enhancing}, MedDG~\citep{liu2022meddg}, and DialoACM~\citep{chen2022benchmark} are datasets of pipeline dialogue systems for automatic diagnosis.
MedDialog~\citep{zeng-etal-2020-meddialog} is a large-scale unannotated dataset, utilized for end-to-end training.

These medical dialogue datasets focus on diagnosis, and ignore consultation. 
Compared with these datasets, MidMed is a medical dialogue dataset for consultation, covering mixed-type dialogues.

\subsection{Mixed-type Dialogue Systems}
Recently, research on the mixed-type dialogue has increased significantly.
These researches fall into two categories: (1) train an all-in-one conversation model by using multiple single-skill conversation datasets, such as persona-chat, task-oriented dialogue, to bind multiple dialogue skills \cite{madotto2020attention,blender2021,Madotto2021TheAA}; (2) collect mixed-type dialog datasets \cite{Kurt2020,smith-etal-2020,liu-etal-2020,Sun2021AddingCT,liu-etal-2021-du,chiu-etal-2022,liu-etal-2022} to train mixed-type dialog models. Those datasets are intended to mix different dialogue skills to meet specific needs, such as recommending movies and songs, and are unable to solve medical consultations.
Compared with them, we collect a mixed-type dialogue corpus, {\dataset}, to facilitate the study of medical consultations.

\section{Dataset Collection}
In this section, we describe the three steps for MidMed construction: (1) Selecting basic diagnosis dialogue data; (2) Constructing annotation guidance; (3) Collecting mixed-type dialogue by crowdsourcing.


\subsection{Selecting Basic Diagnosis Dialogue}
To be close to real-world scenarios, MidMed is constructed based on real diagnosis dialogue dataset MedDialog~\citep{zeng-etal-2020-meddialog}, which is collected from online medical community~\href{https://www.haodf.com/}{haodf.com}.

MedDialog dataset contains 3.4 million Chinese dialogues (consultations) between patients and doctors, covering 29 broad categories of specialties including internal medicine, pediatrics, dentistry, etc., and 172 fine-grained specialties including cardiology, neurology, gastroenterology, urology, etc. 

\textbf{Basic Dialogue Selection}.
For MidMed construction, we recruit twenty medical students, who are experts in four departments, otorhinolaryngology, ophthalmology, skin, and the digestive system department.
To ensure better data quality and construction efficiency, the dialogues only in these four departments are reserved.
Besides, we observe that dialogues with few dialogue utterances are usually of poor quality.
Thus, for high data quality and efficiency of data construction, only those conversations with more than four utterances are kept. 
After the above data processing, there are total 9,000 dialogues obtained.

\textbf{Coarse-grained Privacy Removing}.
Furthermore, for ethical concerns, specific regular expressions for coarse-grained filtering are employed to remove privacy.
To delete patients' privacy, regular expressions, such as `` 
\begin{CJK*}{UTF8}{gbsn}\chuhao{我叫...}\end{CJK*} (My name is ...)'', are designed to delete sentences containing name, gender, and region.
Besides, regular expressions, such as `` \begin{CJK*}{UTF8}{gbsn}\chuhao{陈医生您好...}\end{CJK*} (Hello, doctor Chen, ...)'', are utilized to delete doctors' privacy.

\begin{figure*}[t]
	\small
	\centering
	\includegraphics[width=0.7\linewidth]{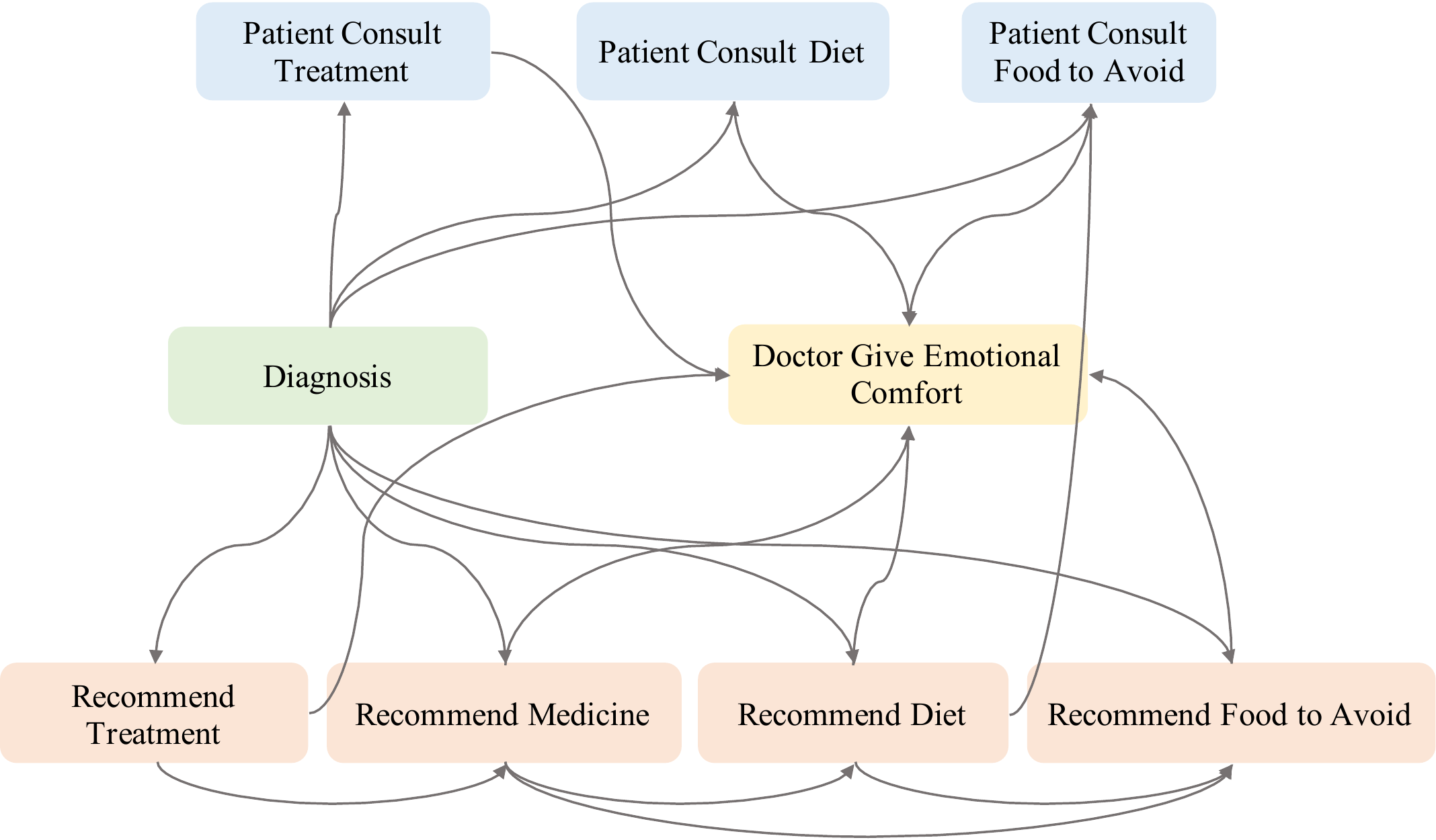}\\
	\caption{The illustration of dialogue topic sequences.}
	\label{figure:goal}
\end{figure*}

\subsection{Constructing Annotation Guidance}
Annotation guidance is designed to instruct annotators for data annotation, including target dialogue topic sequences and reference knowledge.
Specifically, target topic sequences assign topics for each dialogue session.
To support the annotation of each topic, reference knowledge is provided.

\subsubsection{Target Dialogue Topic Sequence}
Due to the complexity of the data annotation, it is of great difficulty to conduct data annotation with only high-level instructions.
Inspired by the work of MultiWOZ~\citep{budzianowski-etal-2018-multiwoz}, we provide a target dialogue topic sequence for each dialogue construction.
The dialogue topic sequences are employed to instruct annotators to annotate the content of specific topics.
As shown in Figure~\ref{figure:sample}, the target dialogue topic sequence is composed of dialogue topics, including \texttt{Patient Self Report}, \texttt{Doctor Inquiry Additional}, \texttt{Doctor Recommend Medicine}, etc.
The whole dialogue topic sequences are shown in Figure~\ref{figure:goal}.
The combination of different topics ensures the diversity of dialogue topic sequences.


\subsubsection{Reference Knowledge}
The knowledge graph stores large-scale knowledge in the form of easy-to-use triples, and it has various applications in all modules of the human-computer dialogue system~\citep{tuan-etal-2019-dykgchat,tuan-etal-2022-towards,yang-etal-2020-graphdialog}. 
Therefore, we incorporate knowledge graphs into medical consultation to provide more accurate interactive questions and answers. 
Specifically, we crawled a large number of web pages from some high-quality medical vertical websites such as 39.net\footnote{http://www.39.net} and then obtained a large amount of triplet knowledge by using information extraction techniques such as entity extraction and relation extraction. 
By using these triples, a large-scale medical knowledge graph is constructed, whose entities include diseases, symptoms, drugs, foods, etc., and relationships include disease-drug relation, disease-food relation, etc. 

To provide reference knowledge for dialogue annotation, we extract a knowledge graph subset for each dialogue.
Specifically, diseases in the whole knowledge graph are mapped with the dialogue with exact string matching.
The disease existing in the medical dialogues are employed as the head entity for select triples from the knowledge graph.
Finally, we extract a knowledge graph subset, which covers four types of entities: disease, symptom, diet, and medicine, with a total of 229,570 triples.

\subsection{Collecting Mixed-type Dialogue}
For data annotation, the trial annotation and the formal annotation are conducted, sequentially.
First, the trial annotation aims to select an annotation team and make the annotation team get familiar with the guide.
Second, the formal annotation is conducted for collecting the whole dataset.

\subsubsection{Trial Annotation}
To ensure the high quality of dialogues, trial annotation is conducted.
In the trial annotation stage, three crowdsourcing teams (about 20 annotators per team) are selected for trial annotation.
There are mainly two advantages. 
(1) Trial annotation helps select a reliable annotation team. 
(2) The trial annotation helps the annotation team get familiar with the annotation task.
Lastly, the team achieving the best performance in the trial annotation is selected for the formal annotation.

\subsubsection{Formal Annotation}
After the trial annotation, the formal annotation is conducted.
In the formal annotation, to ensure data quality, the fine-grained privacy removing, skipping option, and quality audit and re-annotating mechanisms are employed.
To ensure diversity, the mechanism of annotation without target dialogue topic sequences is applied.

\textbf{Overall Annotation}.
In the formal data annotation process, annotators are required to act as doctors and patients in turn.
Annotators construct dialogues based on a given basic diagnosis dialogue, a target dialogue topic sequence, and reference knowledge. 
The annotation progress is conducted as follows.
First, the annotator enters the chat interface to start chatting, and the ``patient'' initiates the conversation.
Second, annotators conduct a dialogue based on the dialogue topic sequence. 
It is important that the information utilized in the dialogue conforms to the reference knowledge.
After successfully mentioning all target topics in sequence, the ``doctor'' ends the conversation.

Furthermore, we introduce the fine-grained privacy removing, the skipping option, quality audit and re-annotating to improve data quality, and introduce the annotation without target dialogue topic sequence mechanism to improve data diversity.

\textbf{Fine-grained Privacy Removing}.
In the data annotation process, for better data quality, annotators are also required to delete privacy that cannot be covered by regular expressions, including gender, age, name, institution name, etc.

\textbf{Skipping Option}.
We observe that there are many basic diagnosis dialogues with low quality.
These bad dialogues may lead to annotated dialogues of low quality.
To alleviate the issue, a skip option is provided to annotators.
Specifically, annotators can choose whether to annotate the given basic diagnosis dialogue or not to the quality of the given dialogue. 
If annotators choose "Skip", they then skip the current dialogue directly and conduct the annotation of the next dialogue. 

To ensure the option is not being overused, we review all the skipped conversations and select high-quality dialogues from the skipped conversations.
Those high-quality dialogues are returned to the annotation process, and the rest low-quality dialogues are abandoned.

\textbf{Quality Audit and Re-annotating}.
To deal with low-quality samples, we introduce the quality audit and re-annotation mechanism. 
Specifically, we review all the annotated samples and pick out low-quality dialogues. 
These low-quality samples are returned to the annotation team for re-annotation.

\textbf{Annotation without Target Dialogue Topic Sequence}.
Though the target dialogue topic sequences lead to good annotation quality, they usually lead to monotonous dialogue structures.
To address the issue, annotators are also allowed to construct the dialogues without following the target dialogue topic sequences.
This option enables annotators to construct more diverse and flexible dialogues based on the basic diagnosis dialogues.
Meanwhile, to prevent this option from being abused, this option is required to be used for no more than ten percent of the whole annotation data.

\begin{table}[t]
\small
\centering
\begin{tabular}{@{}lr@{}}
\toprule
\# of dialogues             &   8,175   \\
~~~~- Otorhinolaryngology      &   1,692   \\
~~~~- Ophthalmology            &   1,443   \\
~~~~- Skin                     &   2,962   \\
~~~~- Digestive System         &   2,078   \\
\# of dialogues w/ goal             &   7,557   \\
\# of dialogues w/o goal            &   752   \\
\midrule
Avg. \# of utterances in a dialogue            &   11.79   \\
~~~~- Otorhinolaryngology      &   12.20   \\
~~~~- Ophthalmology            &   11.02   \\
~~~~- Skin                     &   12.04   \\
~~~~- Digestive System         &   11.64   \\
Max. \# of utterances in a dialogue & 46 \\
Min. \# of utterances in a dialogue & 6 \\ \midrule
\# of tokens                &    1,887,227      \\ 
Avg. \# of tokens in an utterance   & 19.26 \\
Max. \# of tokens in an utterance   & 189 \\
Min. \# of tokens in an utterance   & 2 \\ \bottomrule
\end{tabular}
\caption{Statistics of the MidMed.}
\label{tab:statistics}
\end{table}

\subsection{Dataset Analysis}
\textbf{Data statistics}. Table~\ref{tab:statistics} provides statistics of the MidMed.
There are totally 8,175 dialogues with 11.79 utterances in each dialogue on average.
The longest dialogue contains 46 utterances.
Besides, there are 19.26 tokens in an utterance on average, indicating rich semantic information.

Table~\ref{tab:mixed_type_datasets} lists medical dialogue datasets (MZ~\citep{wei2018task}, DX~\citep{xu2019end}, CMDD~\citep{lin2019enhancing}, MedDG~\citep{liu2022meddg}, MedDialog~\citep{zeng-etal-2020-meddialog}, DialoAMC~\citep{chen2022benchmark} ) and mixed-type dialogue dataset(DuRecDial~\cite{liu-etal-2020}, DodecaDialogue~\cite{Kurt2020}, BlendedSkillTalk~\cite{smith-etal-2020}, ACCENTOR~\cite{Sun2021AddingCT}, DuRecDial 2.0~\cite{liu-etal-2021-du}, SalesBot~\cite{chiu-etal-2022}, DuClarifyDial~\cite{liu-etal-2022}).
MidMed is the first dialogue dataset for consultation, covering five types of dialogues.

\textbf{Data quality}. 
Following~\citep{liu-etal-2020}, for data quality evaluation, we employ human evaluations.
Specifically, we assign ``1'' for dialogues coincident with annotation guidance, and ``0'' for the others.
Then, we conduct a quality evaluation on 100 randomly sampled dialogues. 
Finally, an average score of ``0.90'' is achieved.
The result indicates that the dialogues in the dataset are with high quality.

\begin{figure*}[t]
	\small
	\centering
	\includegraphics[width=\linewidth]{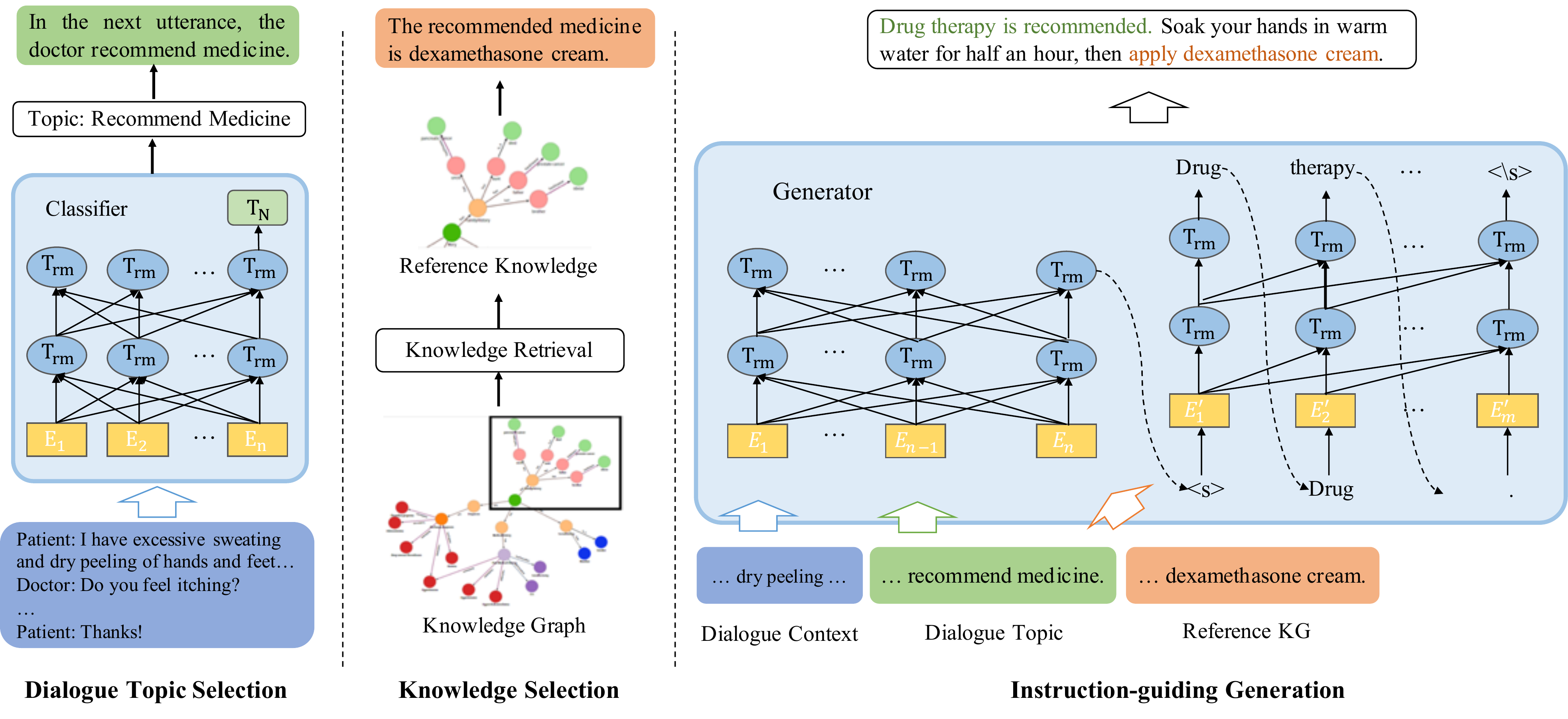}\\
	\caption{The illustration of the proposed InsMed.}
	\label{figure:framework}
\end{figure*}

\section{Method}
During training, a dialogue, with a sequence of utterances between a patient and a doctor, is given.
Then, the dialogue is processed into a set of samples $\{(\bm{s}_{i}, \bm{t}_{i})\} \in \mathcal{D}$, 
where $\bm{t}_{i} $ is $i$-th target doctor response, $\bm{s}_{i}$ is the concatenation of all former utterances before $\bm{t}_{i}$, and $\mathcal{D}$ is the training dataset.
Dialogue generation is formulated as a sequence-to-sequence generation problem, 
which aims to generate $\bm{t}_{i}$ conditioned on $\bm{s}_{i}$.

InsMed has three modules, dialogue topic selecting, reference knowledge selection, and the instruction-guided generation module. 
The dialogue topic prediction and the reference knowledge selection module aim to obtain dialogue topics and reference knowledge, respectively.
Then, for better generation performance, these two types of information are transformed into instructions in natural language.
Finally, instructions are concatenated with context, as the input to generation models.
Next, the above modules are introduced.

\subsection{Dialogue Topic Selection}
The dialogue topic selection module is divided into two stages, the dialogue topic prediction, and the dialogue topic converting.

The dialogue topic prediction aims to predict dialogue topics for the next utterance.
Formally, this task is regarded as a multi-class classification problem.
Specifically, the input of the prediction module is a dialogue context $\bm{s}_{i}$, and the output is the predicted dialogue topics. 
The classification process is formulated,
$$\bm{p}_{i} = f(\bm{s}_{i}),$$
where $f$ is the classification function BERT~\citep{devlin2018bert} and $\bm{p}_{i} \in |\mathcal{R}|^{|\mathcal{C}|}$ is the predicted probability value,  $\mathcal{C}$ is the predefined category set. 
The dialogue topic $a_{i}$ is selected as the predicted dialogue topic if the value of the dimension is the highest probability value in $\bm{p}_{i}$.

Then, in the dialogue topic converting stage, $a_{i}$ is converted into natural language with predefined templates, represented as $\tilde{a}_{i}$.
For example, the predicted topic is \texttt{Recommend Medicine}, and the converted instruction is ``\textit{In the next utterance, the doctor will recommend medicine}''.

\subsection{Reference Knowledge Selection}
The reference knowledge selection module aims to obtain the reference knowledge for model generation, thus guiding models to generate more informative responses.
The module is divided into two parts, knowledge retrieval, and reference knowledge converting.

The knowledge retrieval module aims to retrieve reference knowledge from the whole knowledge graph for response generation. 
An exact string match is utilized for retrieval.
Specifically, the diseases ${\bm{d}}_{i=1}^{~m}$ in the whole knowledge graph are mapped with medical dialogues with exact string matching, where $m$ is the number of diseases.
The disease $d_{i}$ existing in the medical dialogues are regarded as related diseases of the dialogues.
Then, the reference knowledge is obtained by inquiry the knowledge graph with $d_{i}$, 
$\bm{e} = \{tail | \{head, relation, tail\} \in KG, head=\bm{d}_{i}, relation=r\}$,
where $r$ is the slot in the predicted dialogue topic $a_{k}$.
For example, if the dialogue topic is \texttt{Doctor Recommend Medicine}, $r$ is \texttt{Medicine}, and $\bm{e}=\{\rm{bonmopirocin~ointment, dexamethasone}$ $\rm{cream}\}$.

Then, in the reference knowledge converting, $\bm{e}$ is converted into natural language with predefined templates, represented as $\tilde{e}_{i}$.
As the example in Figure~\ref{figure:framework}, the converted knowledge instruction is ``\textit{the recommended medicine is bonmopirocin ointment and dexamethasone cream}''.

\subsection{Instruction-guiding Generation}
The Instruction-guiding generation module aims to generate accurate and informative responses with instructions.

The problem of response generation is formulated as a sequence-to-sequence task~\cite{sutskever2014sequence}.
The input to the generation model is the concatenation of the dialogue context $\bm{s}_{i}$, the predicted dialogue topic instruction $\tilde{a}_{i}$, and the reference knowledge $\tilde{e}_{i}$. 
The output is the doctor's response $\bm{t}_{i}$.

BART~\cite{lewis2019bart} is utilized as the generation model.
Then, the forward calculation process is formulated,
$$\bm{t}_{i} = f_{g}([\bm{s}_{i};\tilde{a}_{i};\tilde{e}_{i}]),$$
where $f_{g}$ represents the generation model BART.

\section{Experiments and Results}
This section introduces
experimental setting,
data and evaluation metrics,
baselines,
automatic evaluations, human evaluations,
and the ablation study.

\subsection{Experimental Setting}
\textbf{Implementation Details}.
For Transformer, the implementation by HuggingFace~\footnote{https://github.com/huggingface/transformers} is utilized,
where the hyperparameters follow the default settings in the original Transformer~\citep{vaswani2017attention}.

For DialoGPT-small~\citep{zhang2018generating}, the layer number, the embedding size, and the context size are set as 10, 768, and 300, respectively.
In layer normalization, the epsilon hyperparameter is set as 1e-5. 
In multi-head self-attention, the number of heads is set as 12. 
The weight parameters are learned with Adam, with the initial learning rate 1.5e-4 and the batch size 32.

For BERT classifier, we use a mini-batch size of 64 and the Adam optimizer with default parameters 
(a fixed learning rate 0.001, $\beta_{1} = 0.9$,  $\beta_{2} = 0.999$, $\epsilon = 1 \times e^{-8}$) \cite{kingma2014adam}. 

For BART, the large version is employed, with the learning rate $2 \times e^{-5}$.
In BART, the BERT encoder and GPT decoder are Transformers with 12 layers and a hidden state size of 768.
The dropout rate is set as 0.1.
The maximum length of input sequences is truncated to 512 and that of output sequences was truncated to 256.

\textbf{Computing Platform}.
Our experiments are conducted on the workstation with an Intel Xeon E5 2.40 GHz CPU, 128 GB memory, an NVIDIA A100 GPU, and CentOS 7.2. 

\subsection{Data and Evaluation Metrics}
We split MidMed into the training set, the validation set, and the test set by randomly sampling 70\%, 10\%, and 20\% data.

\subsubsection{Automatic Evaluation Metrics}
Following~\citet{zeng-etal-2020-meddialog}, four basic automatic evaluation metrics for generation tasks are utilized in this work, including ROUGE~\citep{lin-2004-rouge}, NIST-4~\citep{doddington2002automatic}, BLEU-$n$~\citep{papineni2002bleu} (where $n$ is the size of n-gram), and METEOR~\citep{agarwal2007meteor}.
These metrics all measure the similarity between the generated responses and the ground truth via n-gram matching.

\subsubsection{Human Evaluation Metrics}
Following~\citet{liu-etal-2020}, three human evaluation metrics are utilized in this work, including relevance, informativeness, and human-likeness.

\noindent
\textbf{Relevance} measures fluency, relevancy and logical consistency of each response when given the current goal and global context:
 \begin{itemize}
 \item score 0 (bad): more than two-thirds responses irrelevant or logical contradictory to the given current goal and global context.
\item score 1 (fair): more than one-third responses irrelevant or logical contradictory to the given current goal and global context.
 \item score 2 (good): otherwise.\\
\end{itemize}
\noindent
\textbf{Informativeness} examines how much knowledge (goal topics and topic attributes) is provided in responses:
 \begin{itemize}
 \item score 0 (bad): no knowledge is mentioned at all.
 \item score 1 (fair): only one knowledge triple is mentioned in the response.
 \item score 2 (good): more than one knowledge triple is mentioned in the response.\\
\end{itemize}
\noindent
\textbf{Human-likeness} examines similarity between each generated response with corresponding human response from the perspectives of appropriateness, fluency, and proactivity:
\begin{itemize}
\item score 0 (bad): not like human responses.
\item score 1 (fair): like human responses, but some parts still have deficiencies.
\item score 2 (good): otherwise.
\end{itemize}

\subsection{Baselines}

We carefully select a few strong baselines for comparison.
Specifically, two baselines for mixed-type dialogue generation (BST~\citep{smith-etal-2020}, MGCG~\citep{liu-etal-2020}), a baselines for medical dialogue generation (VRbot~\citep{li2021semi}), two common baselines for medical dialogue (Seq2Seq~\citep{sutskever2014sequence}, DialoGPT~\citep{zhang-etal-2020-dialogpt}), and a baseline for general dialogue generation (BART~\citep{lewis2019bart}) are used in this experiment.
Besides, the proposed model utilizes the same data as these baselines, with domain-specific knowledge.

\textbf{BST}~\cite{smith-etal-2020} is a mixed-type dialogue model that can display many skills, and blend them in a seamless and engaging way.

\textbf{MGCG}~\citep{liu-etal-2020} consists of a goal-planning module and a goal-guided responding module. 
The goal-planning module conducts dialog management to control the dialog flow.
The responding module generates responses for completing each goal.  

\textbf{VRbot}~\citep{li2021semi} introduces both patient state and physician action as latent variables with categorical priors for explicit patient state tracking and physician policy learning, respectively.
A variational Bayesian generative approach is utilized to approximate posterior distributions over patient states and physician actions.

\textbf{Seq2Seq}~\citep{sutskever2014sequence}
\citet{sutskever2014sequence} uses a multilayered Long Short-Term Memory (LSTM) to map the input sequence to a vector of fixed dimensionality, and then another LSTM to decode the target sequence from the vector.

\textbf{DialoGPT}~\citep{zhang-etal-2020-dialogpt} is a large, tunable neural conversational response generation model based on GPT.
DialoGPT is trained on 147M conversation-like exchanges extracted from Reddit comment chains over a period spanning from 2005 through 2017.

\textbf{BART}~\citep{lewis2019bart} is a denoising autoencoder for pretraining sequence-to-sequence models.
It is composed of a BERT encoder  (a bidirectional encoder) and a GPT decoder (a left-to-right decoder).

\begin{table*}[!ht]
\small
\centering
\begin{tabular}{@{}lcccccccc@{}}
\toprule
& ROUGE  & NIST-4 & BLEU & BLEU-1 & BLEU-2 & BLEU-3 & BLEU-4 & METEOR \\ \midrule
BST~\citep{smith-etal-2020}    &  13.64 & 0.81 &  2.88 &  14.01 & 4.89 &  2.15 &  1.02 &  13.81   \\
MGCG~\citep{liu-etal-2020}    &  14.37 & 0.98 &  3.36 &  15.88 & 5.39 &  2.61 &  1.06 &  15.24   \\
VRbot~\citep{li2021semi}       & 23.01 & 1.41       & 4.84    & 22.67       & 8.07       & 3.55       & 1.31       & 18.66       \\
Seq2Seq~\citep{sutskever2014sequence}   & 12.21 &  0.77 & 2.93 & 14.25  &  4.92  &  2.08 &   1.01  &  13.66 \\
DialoGPT~\citep{zhang-etal-2020-dialogpt}    &19.58  &  1.14      & 4.62     & 17.64       &  5.97      & 2.84      & 1.53       &  17.36  \\
BART~\citep{lewis2019bart} &31.63  & 3.12       & 21.95     & 41.36      & 27.26        & 22.04       &  18.87      &  34.74      \\ \midrule
InsMed (Ours)         & \textbf{40.59}           & \textbf{3.30}       & \textbf{23.13}     & \textbf{42.61}       & \textbf{28.46}       & \textbf{23.00}  & \textbf{19.73}       & \textbf{36.45}       \\ 
~~~~w/o Topic   &34.99  & 3.17       & 22.41     & 42.03      & 27.97        & 22.60       &  19.26      &  35.26  \\
~~~~w/o KG   & 32.22  & 3.14    & 22.19  & 41.13  & 27.21 & 22.12 & 18.91  & 34.80  \\
\bottomrule
\end{tabular}
\caption{Automatic evaluation results of five baseline models and InsMed, on eight evaluation metrics. Values of ROUGE, BLEU, and METEOR are expressed as percentages (\%).}
\label{tab:results}
\end{table*}

\begin{table*}[!ht]
\centering
\small
\begin{tabular}{@{}lcccccccc@{}}
\toprule
& BST & MGCG & VRbot & Seq2Seq & DialoGPT & BART & InsMed (Ours) & Groundtruth \\ \midrule
Relevance    &  0.33   &  0.39  & 0.42    & 0.27   & 0.50  & 1.32 & 1.42 & 1.98  \\
Informativeness &  0.29 &  0.31  & 0.36  & 0.24 & 0.46  & 1.30  & 1.54  & 2.00 \\
Human-likeness  &  0.37 & 0.41  & 0.48  & 0.32 & 0.70  & 1.68   & 1.88 &  2.00 \\ \bottomrule
\end{tabular}
\caption{Human evaluation results of six baseline models, InsMed, and Groundtruth, on three aspects, including relevance, informativeness, and human-likeness. Scores of ``0'', ``1'', and ``2'' are assigned to each dialogue, where ``0'' represents bad samples and ``2'' represents good samples. The average scores are reported.}
\label{tab:human_evaluation}
\end{table*}

\subsection{Automatic Evaluation}
The results on automatic evaluation metrics are shown in Table~\ref{tab:results}.
InsMed is compared with the other five generation models on various evaluation metrics.
The results show the following conclusions.

First, BART (large) is much better than other baseline generation models.
The reason may be that BART (large) is much more powerful than other generation models, with more parameters and more training data.

Second, InsMed achieves state-of-the-art performance on almost all metrics.
This demonstrates that instructions help BART to generate more accurate responses.

\subsection{Human Evaluation}
Table~\ref{tab:human_evaluation} shows the human evaluation results on the test set of MidMed.

First, comparing BART, InsMed with other baselines, the results demonstrate that pre-training on large-scale data improves relevance, informativeness, and human-likeness.
The reason may be that pre-training on large-scale data provides a large amount of common language knowledge.

Second, comparing InsMed with BART, the results show that InsMed performs better than BART, especially in relevance and informativeness.
The reason may be that instructions in InsMed provide specific targets for generation, leading to a more relevant and informative response generation.

\subsection{Ablation Study}
Table~\ref{tab:results} shows the ablation results, where ``w/o Topic'' means removing dialogue topic instructions from the InsMed and ``w/o KG'' means removing reference knowledge instructions from the InsMed.
Results show that reducing any module of MidMed leads to poor results.
This illustrates the effectiveness of each module of the InsMed.

\section{Conclusion}
This work identified the challenge of helping patients clarify their goals through medical consultations.
To address this challenge, this work proposed a novel task, medical consultation over mixed-type dialogue, and collected a new Chinese human-to-human mixed-type dialogue dataset, in which each session has rich variability of dialogue types with natural topic transitions.
To facilitate further research, we conducted benchmarking experiments on MidMed for end-to-end dialogue generation and proposed an instruction-guiding medical dialogue generation framework InsMed.
Experimental results show the effectiveness of InsMed.
In the future, we will investigate the possibility of cross-departments (e.g. dermatology and endocrinology) medical consultation at low cost.

\section{Limitation}
InsMed is built based on the large-scale pre-training model BART, which requires high computing resources.
Besides, the data currently only covers four departments, limiting the usage scenarios of the data.

\section{Ethical Statement}
We make sure that MidMed is collected in a manner that is consistent with the terms of use of any sources and the intellectual property and privacy rights of the original authors of the texts. 
And crowd workers were treated fairly. 
This includes, but is not limited to, compensating them fairly, ensuring that they were able to give informed consent, and ensuring that they were voluntary participants who were aware of any risks of harm associated with their participation.

\section{Acknowledgements}
Thanks for the insightful comments from reviewers.
This work is supported by the Shanghai Artificial Intelligence Laboratory.

\bibliography{xmshi}
\bibliographystyle{acl_natbib}

\end{document}